\documentclass[letterpaper, 10 pt, conference]{ieeeconf}

\usepackage{cite}
\usepackage{amsmath,amssymb,amsfonts}
\usepackage{graphicx}
\usepackage[hidelinks]{hyperref} 
\usepackage{textcomp}
\usepackage{tabularx}
\usepackage{amsmath}
\usepackage{balance}
\usepackage{multirow}
\usepackage{dblfloatfix}
\usepackage{siunitx}
\usepackage{booktabs}
\usepackage{longtable}
\usepackage{cite}
\usepackage{color,colortbl}
\usepackage{algpseudocode}
\usepackage{algorithm}
\definecolor{Gray2}{gray}{0.9}
\definecolor{Gray}{gray}{0.7}

\usepackage{xcolor}
\usepackage{subcaption}

\IEEEoverridecommandlockouts 

\begin{document}

\title{\LARGE \bf AgilePilot: DRL-Based Drone Agent for Real-Time Motion Planning in Dynamic Environments by Leveraging Object Detection\\
}

\author{Roohan Ahmed Khan, Valerii Serpiva, Demetros Aschalew, Aleksey Fedoseev, and Dzmitry Tsetserukou
\thanks{The authors are with the Intelligent Space Robotics Laboratory, Center for Digital Engineering, Skolkovo Institute of Science and Technology, Moscow, Russia. 
\tt \{roohan.khan, valerii.serpiva, demetros.aschu aleksey.fedoseev, d.tsetserukou\}@skoltech.ru}
}

\maketitle

\begin{abstract}
Autonomous drone navigation in dynamic environments remains a critical challenge, especially when dealing with unpredictable scenarios including fast-moving objects with rapidly changing goal positions. While traditional  planners and classical optimization methods have been extensively used to address this dynamic problem, they often face real-time, unpredictable changes that ultimately lead to suboptimal performance in terms of adaptiveness and real-time decision-making. In this work, we propose a novel motion planner, AgilePilot, based on deep reinforcement learning (DRL) that is trained in dynamic conditions, coupled with real-time computer vision (CV) for object detections during flight. The training-to-deployment framework bridges the Sim2Real gap, leveraging sophisticated reward structures that promote both safety and agility depending upon environmental conditions. 

The system can rapidly adapt to changing environments while achieving a maximum allowable speed of $3.0$ m/s in real-world scenarios. In comparison, our approach outperforms classical algorithms such as the Artificial Potential Field (APF)-based motion planner by $3$ times, both in performance and tracking accuracy of dynamic targets by using velocity predictions while exhibiting $90\%$ a success rate in $75$ conducted experiments. This work highlights the effectiveness of DRL in tackling real-time dynamic navigation challenges, offering intelligent safety and agility.

\end{abstract}
{Keywords: Drone Navigation, Dynamic Environments, Motion Planning, Deep Reinforcement Learning, Safety and Agility, Computer Vision, Robotics.}


\section{Introduction}
In recent years, autonomous drone navigation has gained significant interest, especially in the context of dynamic environments. Conventional motion planning methods, such as nonlinear Model Predictive Control (NMPC) and graph-based search algorithms, have demonstrated efficiency in structured and controlled environments but lack the capability to adapt to rapidly changing conditions in real-time. Therefore, DRL has emerged as a viable alternative, offering fast decision-making and adaptability that improve autonomous navigation. In recent work, Song et al. \cite{DRLracing10.1109/IROS51168.2021.9636053} have explored near-optimal trajectory generation in near time using DRL methods, enabling adaptation to dynamic track configurations while achieving remarkable speeds of up to 60 \text{km/h}. Another study \cite{Beautybeastinproceedings} by Kaufmann et al. utilizes Model Predictive Control (MPC) to navigate quickly through the track, where a convolutional network and an Extended Kalman Filter (EKF) predict the poses of the closest gates along with uncertainty. Moreover, another breakthrough study \cite{ChampiondroneKaufmann2023} presents a SWIFT system that competes with human drone pilots in drone racing by using the Proximal Policy Optimization (PPO) algorithm to train a model that generates low-level control commands for the quadrotor. Despite these studies having developed agile systems for drones, they lack the ability to effectively avoid obstacles in real time and adapt to a highly dynamically changing environment. In addition, there are also limitations to their ability to balance safety and agility when the environment becomes highly unpredictable.

\begin{figure}[t]
\centering
\includegraphics[width=1.0\linewidth]{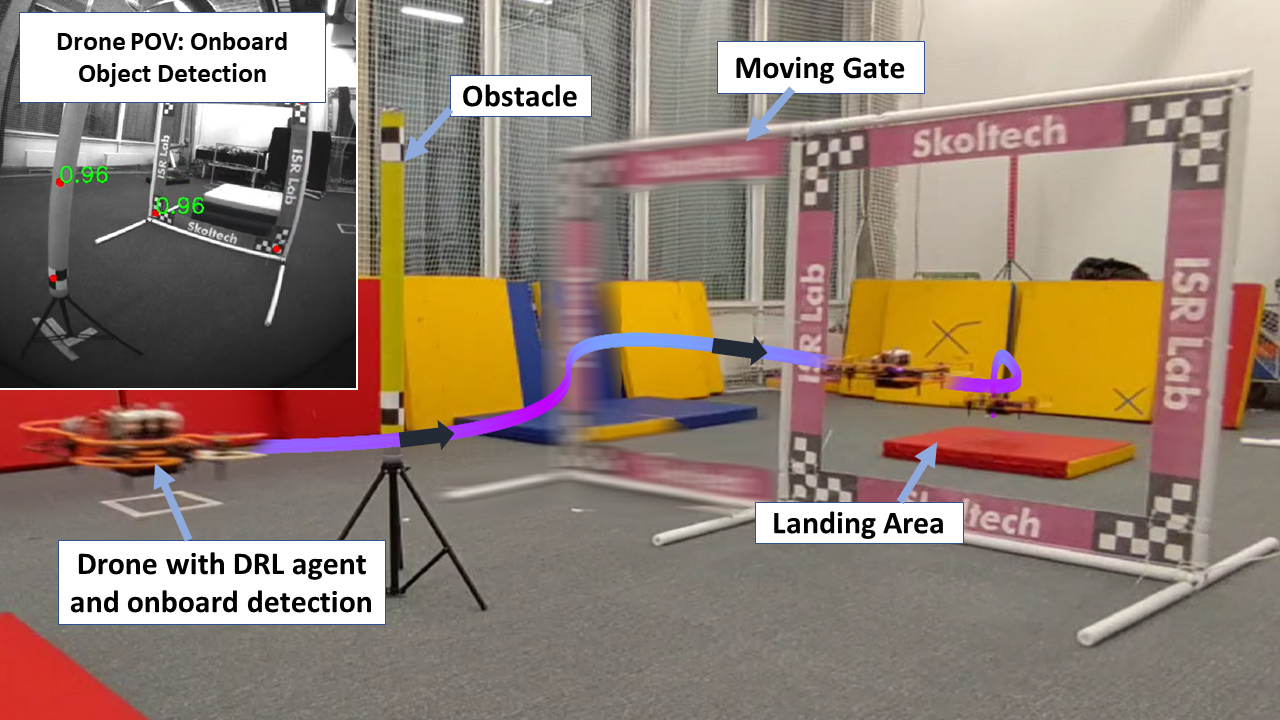} 
\caption{AgilePilot technology performs deep reinforcement learning-based motion planning to avoid obstacles and navigate through moving gates.}
\label{title}
\vspace{-0.5cm}
\end{figure}

This research introduces a novel motion planner, AgilePilot, which leverages DRL to predict drone velocities for navigating dynamic environments with moving objects. The velocity prediction is situation-aware, reaching up to 3.0 m/s in simpler terrain while slowing down in complex scenarios involving obstacles and moving objects. The proposed methodology involves a unique framework to train models in a simulation environment customized to achieve the level of accuracy required for Sim2Real transfer. The training is performed using actor-critic neural network architecture and uses a unique reward structure with effective randomization during episodes to make the application of trained models generalized and easy to implement. To track movable objects, the system applies YOLO-based object detection combined with state estimation using IPPE PnP and an EKF for robust pose estimation. 

The methods used in this research revolutionize the motion planning of drones in dynamic environments by effectively adapting the velocity vector according to the complexity of terrain while taking into account uncertainty in the objects movements. The broad applicability of technology is ensured by the pipeline used for training and testing of trained models in unknown environments. 


\section{Related Works}

Due to the overwhelming complexities of drone navigation in dynamic environments, many studies have been trying to solve the general applicability of algorithms either by using machine learning or classical methods. For example, in study \cite{Sim2real10553074}, the policy was learned with high uncertainty and tested in a noisy real environment. The results showed effective drone navigation in reality, but the method is applicable in 2D space only and not tested in dynamic conditions. 

In study \cite{NavigateObstacleDRL9081749}, DRL is utilized to avoid stationary and moving obstacles, with observation consisting of images and several scalars with Joint Neural Network (JNN). Liu et al. \cite{searchtocontrolreinforcementlearningbased} use a motion planning framework that integrates visibility path searching to generate collision-free paths and RL to generate low-level motion commands. Moreover, CPU-based trained DRL for UAV-to-UAV tracking is introduced in \cite{PPOtrackingTAN2023101497}. The model was trained using PPO and showed considerable results. Another PPO-based learned RL agent in study \cite{LongtermDRL9259811} was used in a racing environment as a path planner and utilized a conventional proportional–integral–derivative (PID) controller for controlling the motion. Moreover, Song et al. \cite{SmoothTrajectoryCollision10069287} proposed a system that generates smooth, collision-free trajectories while taking care of an unseen environment by incorporating vision with DRL. While these mentioned studies show promising capabilities for motion generation and obstacle avoidance, they have predominantly been tested in simulation environments only with less challenging scenarios.

Furthermore, in study \cite{flyingswarm} highly efficient path planner was devised for cluttered environments incorporating some extent of dynamical feasibility by adjusting time allocation based upon spatial-temporal joint optimization. Moreover, study \cite{searchbasedapp13042244} introduced a search-based method by using a multi-stage training approach and testing the agent on a real air-ground unmanned system. In \cite{denseurbanen17112762}, DRL was trained for random obstacles in a dense urban environment by utilizing a Double Deep-Q network (DDQP) while enhancing the training stability of algorithms. In addition, study \cite{mapparams} improves safe navigation in obstacle avoidance by leveraging map parameterization and low-cost planning, where the results are validated in real-life experiments. Peter et al. in study \cite{landeraiadaptivelandingbehavior} trained models using the Twin-Delayed Deep Deterministic (TD3) policy gradient algorithm, and real-life experiments showed remarkable landing performance of drones for dynamically moving platforms in the presence of downwash and wind turbulence. The extension of these works using PPO has also been successfully implemented for multi-agents \cite{marlanderlocalpathplanning}. However, these approaches focus purely on path planning without considering obstacles and have only been tested on tiny-sized drones, therefore, overall restricting the high velocity potential of medium-sized drones. Lastly, another study \cite{omnirace6dhandpose} introduced OmniRace, a control interface based on drone velocity manipulation for drones of all sizes. However, its motion planning depends entirely on human input, which limits both navigation accuracy and autonomy.

In order to address the mentioned gaps, this research presents a novel DRL-based motion planning framework that enables velocity-based trajectory generation for any size drone due to its nature of being a model-free learning agent. Unlike previous studies that rely on static environments, our approach is designed for dynamic scenarios, ensuring robust adaptability to rapidly changing surroundings. The training environment used is Gym PyBullet \cite{panerati2021learning} and is carefully customized and tuned to incorporate real-world complexities for a seamless transition from simulation to reality. Furthermore, this framework accounts for dynamically evolving obstacles, allowing the drone to make intelligent decisions in real time while balancing agility and safety as required. 

\section{Methodology}

\begin{figure}[t]
\centering
\includegraphics[width=1.0\linewidth]{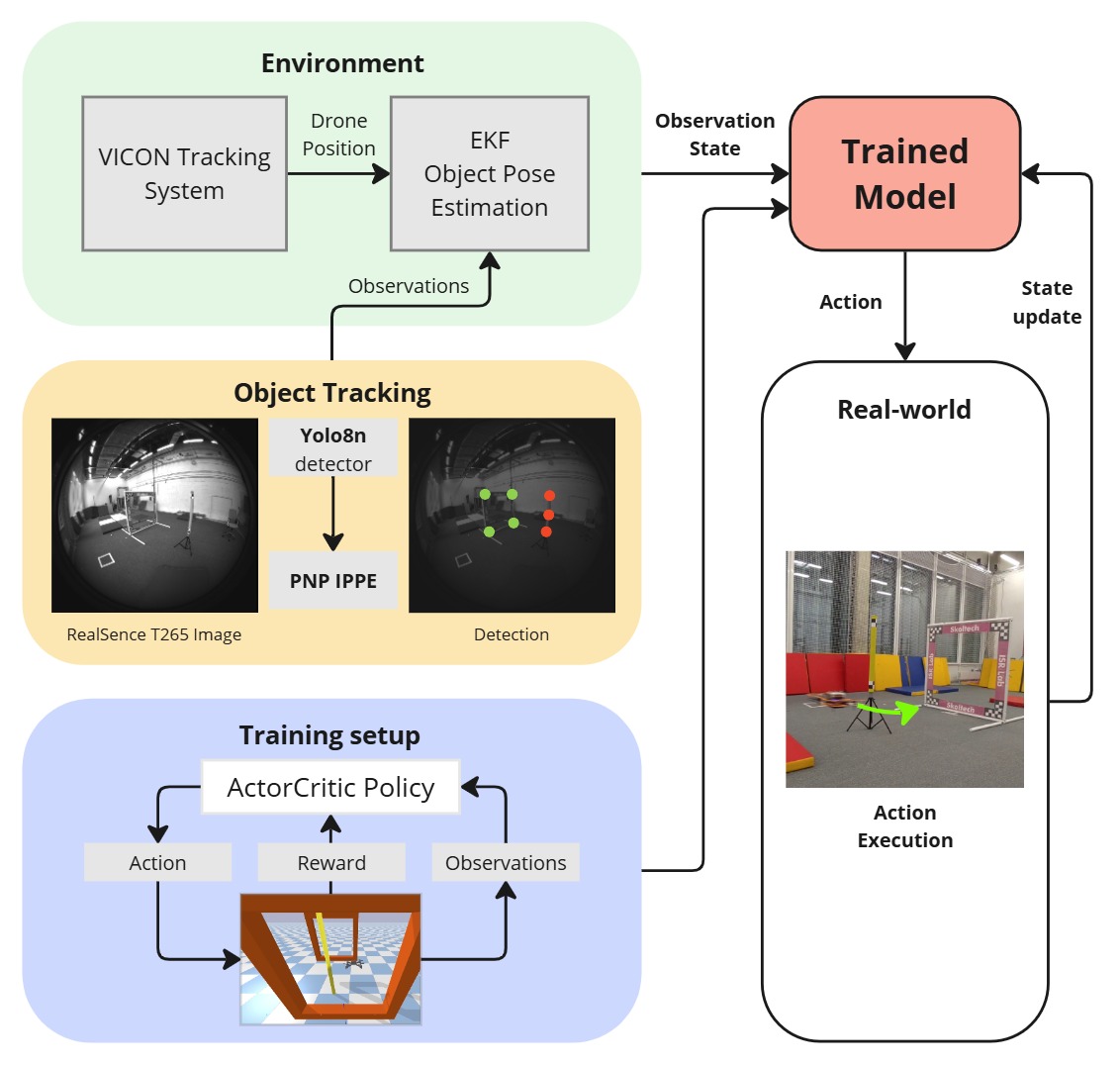} 
\caption{AgilePilot consists of a position estimation module for the agent and dynamic objects in the environment, along with a control policy that maps state observations to control commands.}
\label{methodology}
\vspace{-0.3cm}
\end{figure}

This section explains the methods used for developing the AgilePilot pipeline. Gym PyBullet is used for simulation and physics in training with PPO, and models are deployed in simulated and real-life environments. The overall pipeline of our system is shown in Fig. \ref{methodology}. 

\subsection{Simulation Environment}

\begin{figure}[t]
\centering
\includegraphics[width=0.8\linewidth]{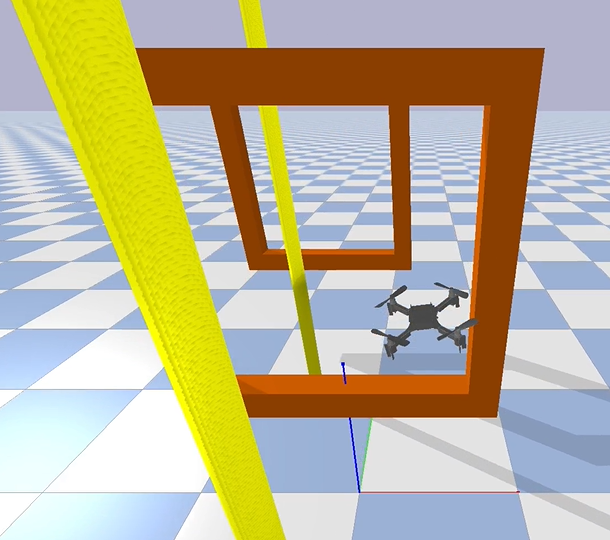} 
\caption{Gym PyBullet custom simulation environment.}
\label{simulation}
\vspace{-0.3cm}
\end{figure}

The policy is trained using an actor-critic network utilizing the PPO algorithm in a customized simulated environment. The position estimation uses a VICON tracking system to provide the drone's state, while object detection is performed using a YOLO neural network for gate corner detection and obstacle keypoint extraction. The keypoints are processed using IPPE PnP to estimate local positions, which are then mapped to a 3D pose in the global coordinate system. The simulation setup is developed using the Gym environment and the PyBullet physics engine that allows realistic aerodynamics and accurate quadcopter mathematical modeling to emulate drones. We have integrated a custom midsized drone with a custom PID controller in simulation. The inertial and controller parameters are tuned to the level of realistic response in real environments. Moreover, custom objects such as gates and obstacles are also built, as shown in the simulation in Fig. \ref{simulation}. 

In each step, the object's movement is governed by a random velocity drawn from a uniform probability distribution:


\[
\Delta \mathbf{p} \sim \mathcal{U}(-v_{\text{max}}, v_{\text{max}}),
\]
where \( \Delta \mathbf{p} = [\Delta x, \Delta y] \) is the change in position in the $x$ and $y$ directions, and \( \mathcal{U}(-v_{\text{max}}, v_{\text{max}}) \) is the uniform distribution over the range \( -v_{\text{max}} \) to \( v_{\text{max}} \).

Lastly, the PID controller allows the drone to reach velocities within a range of $\pm3$ m/s in $xy$ and $\pm2$ in $z$ direction, whereas rotation angles can range from $-\pi$ to $\pi$ radians, by mapping desired commands to motor RPMs and subsequently to desired forces and torques by using an accurate motor dynamics model. 

\subsection{Deep Reinforcement Learning}
\subsubsection{Neural Network Architecture}

\begin{figure}[t]
\centering
\includegraphics[width=1.0\linewidth]{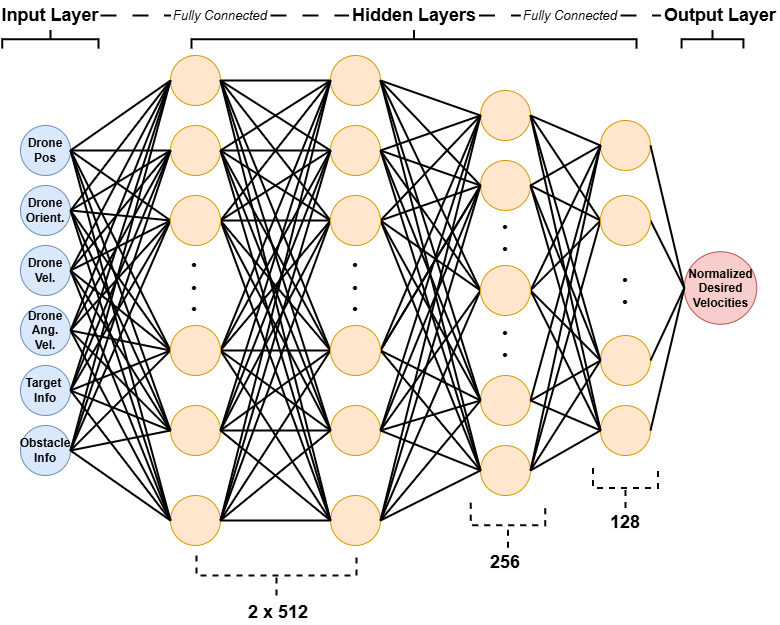} 
\caption{Neural network architecture representing the input layer, hidden layers, and output layer of our system.}
\label{nn}
\vspace{-0.3cm}
\end{figure}

Our neural network follows an actor-critic shared architecture Fig. \ref{nn}. The input layer consists of the following observations:

\[
\mathbf{\hat{o}}_t = \left[ \mathbf{x}_{\text{drone}}, \mathbf{\theta}_{\text{drone}}, \mathbf{v}_{\text{drone}}, \mathbf{\omega}_{\text{drone}}, \mathbf{d}_{\text{target}}, \Delta \mathbf{r}_{\text{obs}} \right].
\]
where $\mathbf{x}_{\text{drone}}$ represents the drone's position, consisting of its coordinates, $\mathbf{\theta}_{\text{drone}}$ denotes the drone's orientation, $\mathbf{v}_{\text{drone}}$ is the linear velocity of the drone, and $\mathbf{\omega}_{\text{drone}}$ indicates the drone's angular velocity. The target's state is represented by $\mathbf{d}_{\text{target}}$, which includes the target's 3D position, its size, and orientation. Lastly, $\Delta \mathbf{r}_{\text{obs}}$ denotes the relative position of an obstacle as well as the obstacle's size.

The output layer of the actor represents the action space, which consists of the following four values:

\[
\mathbf{\hat{a}}_{t+1} = \left[ v_x, v_y, v_z, v_{\text{max}} \right],
\]
where the desired velocity components of the drone are represented by $v_x$, $v_y$, and $v_z$, and $v_{\text{max}}$ denotes the maximum velocity the drone can achieve.

The neural network is designed with ReLU activation functions and uses fully connected (FC) layers with the following structure:
\[
\text{ReLU}(\text{FC512} \times 2) \to \text{ReLU}(\text{FC256}) \to \text{ReLU}(\text{FC128}).
\]

The output layer is essentially a four-dimensional velocity vector, which is controlled by a low-level PID controller. The velocity vector is tunable by some scale factor for implementation in the real-world drone control system. Moreover, the desired yaw is determined indirectly by projecting the xy velocity vector from the action space.

\subsubsection{Reward Structure}

\begin{figure}[t]
\centering
\includegraphics[width=0.8\linewidth]{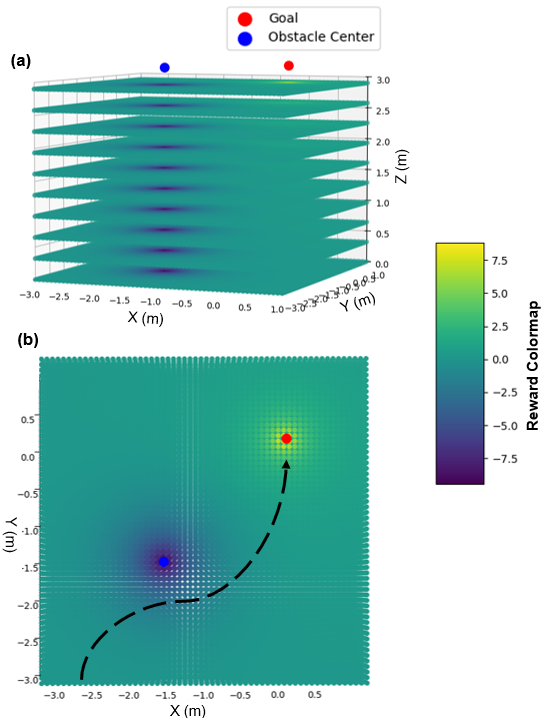} 
\caption{Colormap visualizes reward values for drone positions in the presence of an obstacle and a goal in a defined area, i.e., (a) shows reward structure in 3D space where the obstacle is considered as infinite height, and (b) shows a top view in the XY plane with the predicted path of the drone that avoids the obstacle and reaches the goal.}
\label{reward}
\vspace{-0.6cm}
\end{figure}

The reward structure for the drone is designed to incentivize efficient goal traversal while avoiding obstacles Fig. \ref{reward}.

The agent receives a reward based on its distance to the goal. The reward increases substantially when the drone is in close proximity to the goal, which allows stable target location tracking in 3D space.

\[
R_{\text{proximity}} = \frac{1}{d_{\text{goal}} + c_{p}},
\]
where \( d_{\text{goal}} \) is the Euclidean distance between the drone's position and the goal, and \(c_{p}\) is the constant to avoid extremely high reward values.

Moreover, the agent is penalized for getting closer to the obstacle where the obstacles are modelled as infinite-height objects. The penalty is based on the relative distance to the obstacle and its size, using an exponential function:

\[
R_{\text{obstacle}} = -c_{o} \cdot \exp\left( - \frac{d_{\text{obstacle}}}{r_{\text{safety}}} \right),
\]
where \( d_{\text{obstacle}} \) equals \( \| \Delta \mathbf{r}_{\text{obs}_{xyz}} \| \), \(c_{o}\) is the penalty scaling constant and \( r_{\text{safety}} \) is the safety region around the obstacle.

Lastly, penalties are applied either if the drone collides with an object, with a large constant, $c_{penal}$ or if the drone's velocity is very high within the safety region to promote safe navigation behavior.

\[
R_{\text{collision}} = 
\begin{cases} 
-c_{penal} & \text{if collision} \\
0 & \text{else}
\end{cases},
\]

\[
R_{\text{velocity}} = 
\begin{cases} 
-c_v \cdot \| \mathbf{v} \|^2 \quad & \text{if } d_{\text{obstacle}} < r_{\text{safety}} \\
0 & \text{else}
\end{cases},
\]
where \( \| \mathbf{v} \| \) is the Euclidean norm of the velocity vector, and $c_{v}$ is the velocity penalty constant.

Finally, the total reward \( R_{total} \) is computed as:
\[
R_{total} = R_{\text{proximity}} + R_{\text{obstacle}} + R_{\text{collision}} + R_{\text{velocity}}.
\]

\subsubsection{Training}

The drone is randomized within the following bounds:
\begin{itemize}
    \item \(x, y \in [-4.0, 4.0]\) (in the \(xy\)-plane)
    \item \(z \in [0.3, 4.0]\) (in the \(z\)-axis)
    \item $\mathbf{\theta}_{\text{drone}}$ within \([- \frac{\pi}{2}, \frac{\pi}{2}]\).
\end{itemize}

The obstacle position is randomized based on two components: longitudinal and lateral offsets. The longitudinal offset is calculated along the line connecting the drone and the target, while the lateral offset is perpendicular to this line. \(\vec{L}\) is the vector from the drone to the target, calculated as:
\[
\vec{L} = \text{$\mathbf{d}_{\text{target}_{xyz}}$} - \text{ $\mathbf{x}_{\text{drone}}$}.
\]
The longitudinal offset \(d_{\text{long}}\) is given by:
\[
\vec{O}_{\text{long}} = d_{\text{long}} \cdot \vec{L}.
\]
The lateral offset is computed as the cross product of \(\vec{L}\) with the unit vector \(\vec{Z}\): 
\[
\vec{L}_{\text{lat}} = \frac{\vec{L} \times \vec{Z}}{\|\vec{L} \times \vec{Z}\|}.
\]
The obstacle’s position is then determined by combining the longitudinal and lateral offsets with the drone's initial position in the episode.
The \(z\)-coordinate of the obstacle is randomized within a range, typically around \(z \in [0, 2]\).

\begin{table}[htbp]
\centering
\caption{Training Parameters for DRL}
\begin{tabular}{|c|c|}
\hline
\textbf{Parameter} & \textbf{Value} \\
\hline
Algorithm & PPO (Proximal Policy Optimization)  \\
\hline
Total Steps & 25,000,000 \\
\hline
Number of Environments & 8 \\
\hline
Batch Size & 256 \\
\hline
Number of Steps & 2048 \\
\hline
Entropy Coefficient & 0.01 \\
\hline
Discount Factor & 0.99 \\
\hline
Clip Range & 0.2 \\
\hline
Activation Function & ReLU \\
\hline
\end{tabular}
\label{hyperparams}
\end{table}

Moreover, Table ~\ref{hyperparams} summarizes the key parameters used in the reinforcement learning algorithm within the training. The optimization algorithm used for training is Adam, which is a widely used optimizer in reinforcement learning. The learning rate is set to \( 1 \times 10^{-4} \).

The feature extractor (policy network) processes observations and provides the necessary features for action prediction. For PPO, the policy architecture is based on anactor-critic, which integrates both the actor (policy) and critic (value function) into a single neural network. The output layer provides a distribution of actions based on the processed input.

\begin{figure}[t]
\centering
\includegraphics[width=1.0\linewidth]{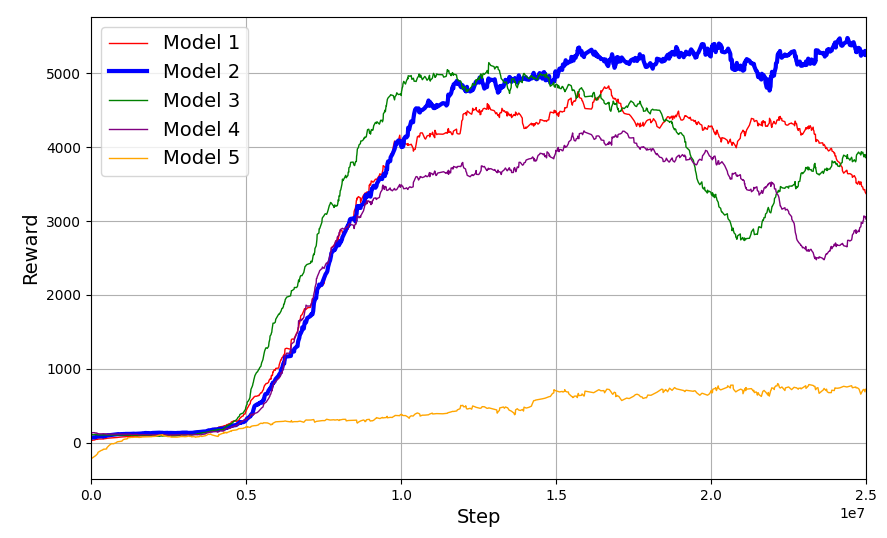} 
\caption{The mean reward of five trained models with different parameters shows that reward increases with time before converging.}
\label{training_reward}
\vspace{-0.3cm}
\end{figure}

\begin{figure}[t]
\centering
\includegraphics[width=1.0\linewidth]{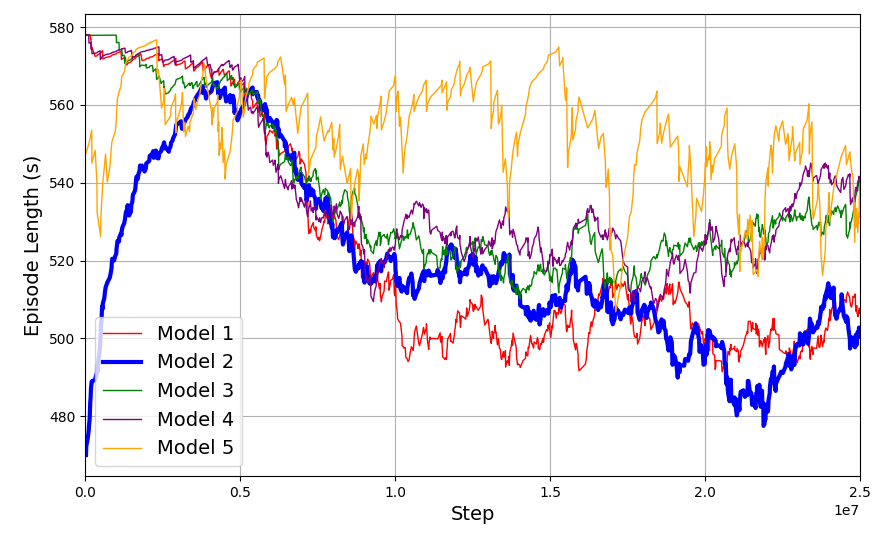} 
\caption{The mean episode length of five trained models with different parameters shows decreasing trend with time that indicates model is converging to the best learned policy.}
\label{training_ep}
\vspace{-0.5cm}
\end{figure}

The model’s performance is evaluated after training by running it in the environment and computing the mean and standard deviation of the rewards over multiple episodes. Fig. \ref{training_reward} and Fig. \ref{training_ep} show the mean reward and mean episode length, respectively, of several trained models with different sets of hyperparameters. Model 2 with hyperparameters listed in Table. \ref{hyperparams} was selected for the experiment due to its better convergence with the highest reward and minimum episode length, indicating that the policy has learned well.

\subsection{Computer Vision System}


\subsubsection{Object Detection}

The object detection system receives an image from the wide-angle Intel RealSense T265 camera with 30 fps as input and outputs an estimate of the spatial position of the gates and obstacles relative to the camera. The camera captures a frame, which is then processed by the YOLOv8n Pose neural detector. The detector identifies four key points that define the gate plane with a size of 1.5 x 1.5 m, as well as three key points for the cylindrical obstacle with a length of 1.0 m and a diameter of 0.1 m. To train the detection model, a custom dataset was collected, consisting of 2475 annotated instances of racing gates and 3015 instances of obstacles. 4471 images were generated from the original data set using specialized augmentation techniques. The final dataset was split into training and validation sets in an 80:20 ratio. The YOLO model was trained for 500 epochs with a batch size of 32 with an image resolution of 424x400 px. Data augmentation was applied, including horizontal flipping with a probability of 0.5, random rotation up to 10 degrees, and color adjustments in the HSV space with the following parameters: hue 0.015, saturation 0.2, and value 0.2. The inference time of the model is 40 ms, demonstrating the system's capability for real-time processing.

\subsubsection{Position estimation}

\begin{figure}[t]
\centering
\includegraphics[width=0.8\linewidth]{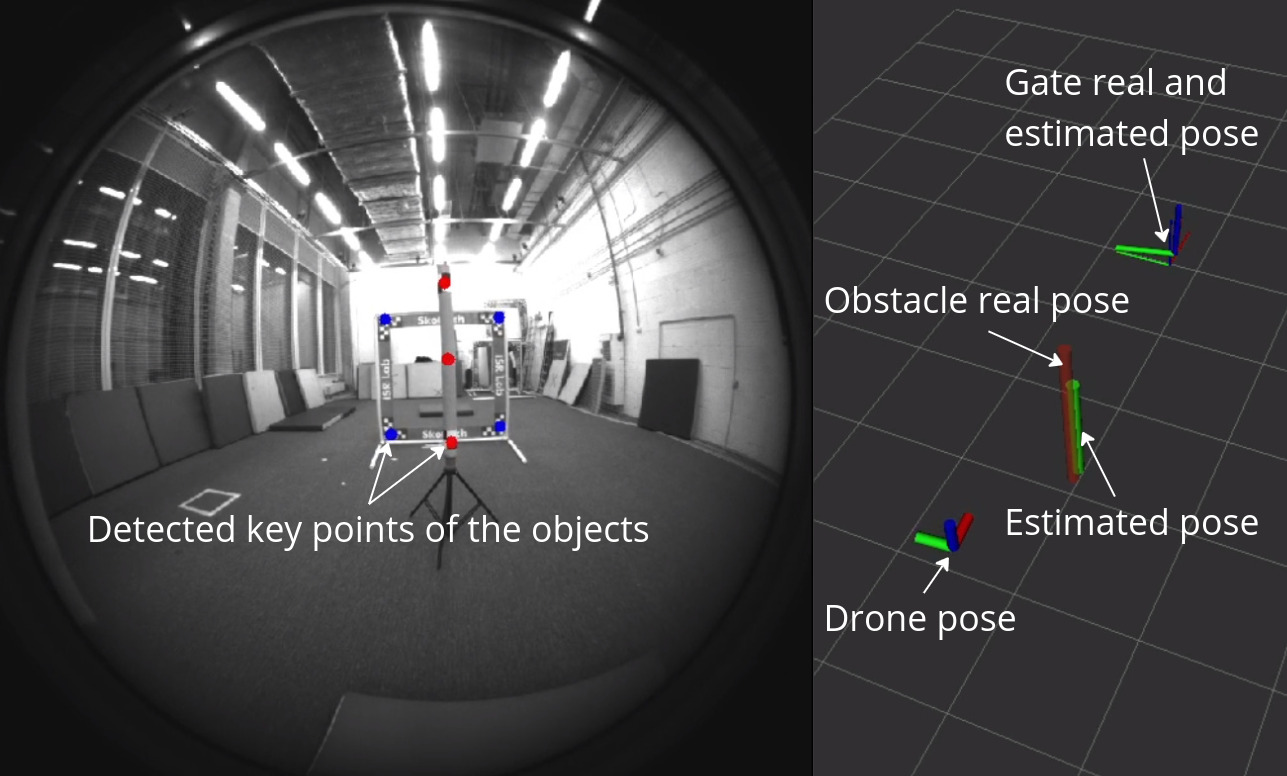} 
\caption{Detected key points on the frame and the visualization of the estimated poses of the objects in RVIZ.}
\label{detection_result}
\vspace{-0.4cm}
\end{figure}

The Infinitesimal Plane-based Pose (IPPE) method facilitates the estimation of the spatial position and orientation of racing gates and obstacles relative to the drone's camera. This is accomplished using the 2D coordinates of the object's corner points, the 3D coordinates of these corners within the object's local coordinate system and the intrinsic camera parameters. The method allows for the precise determination of the pose and orientation of the object in space. The EKF is employed to filter noisy position measurements of multiple objects, enabling accurate tracking by associating observations based on their spatial proximity. The state vector for gates is represented as $\mathbf{x} = \begin{bmatrix} x & y & z & \psi \end{bmatrix}^T$ and for cylindrical obstacle $\mathbf{x} = \begin{bmatrix} x & y \end{bmatrix}^T$  represent the position and orientation of an object, where: $x, y, z$ are the position coordinates and $\psi$ is the yaw angle (orientation). The results of the working detection and pose estimation system are presented in Fig. \ref{detection_result}.


\section{Comparative Analysis}
In this analysis, we compare the performance of our DRL-based motion planner with the well-known Artificial Potential Field (APF) motion planner due to its effectiveness in dynamic environments for UAVs \cite{adaptiveapf}, \cite{uavapf}. To perform this comparison, we conducted five simulation cases to assess different dynamic conditions. The general configuration for each case includes two obstacles and one gate that the drone must pass through to reach the target location. 

\subsection{Simulation Cases}
The five simulation cases are as follows:

\begin{itemize}
    \item \textbf{Case 1:} Gate moving at 0.3 m/s with obstacle placed to exploit APF local minima.
    \item \textbf{Case 2:} Gate moving at 0.3 m/s.
    \item \textbf{Case 3:} Gate moving at 0.3 m/s, varying target height.
    \item \textbf{Case 4:} Gate and obstacle moving at 0.3 m/s, varying target height.
    \item \textbf{Case 5:} Gate moving at 0.6 m/s.
\end{itemize}

Each case is performed 15 times with distinctive initial drone and object positions to record the number of failures, where failure exists whenever the drone collides with the object in the environment.

\subsection{Performance Metrics}
The performance metrics are given:

\begin{itemize}
    \item \textbf{Success Rate:} The success rate is calculated as:
    \[
    \text{Success Rate (\%)} = (1 - \frac{\text{Number of Failures}}{15}) \cdot 100.
    \]
    \item \textbf{Tracking Precision:} The precision to track the moving gate and target position.
    \item \textbf{Time to Complete Experiment:} The total time taken for the drone to complete the full task in each case.
\end{itemize}

\begin{table}[h!]
\centering
\caption{Success Rate Comparison of DRL Agent vs APF}
\begin{tabular}{|c|c|c|}
\hline
\textbf{Case} & \textbf{DRL Agent} & \textbf{APF} \\
\hline
\textbf{1} & $100\%$ & $0\%$ \\
\hline
\textbf{2} & $100\%$ & $80\%$ \\
\hline
\textbf{3} & $90\%$ & $60\%$  \\
\hline
\textbf{4} & $80\%$ & $40\%$ \\
\hline
\textbf{5} & $80\%$ & $20\%$ \\
\hline
\end{tabular}
\label{successrate}
\end{table}

Table \ref{successrate} showcases the average $90\%$ success rate out of a total of 75 experiments by the DRL agent as compared to the APF motion planner, where the APF completely failed to navigate through the local minima situation in Case 1 and struggled to complete its task for the highly dynamically moving environment in Case 4 and Case 5.

\begin{table}[h]
\centering
\caption{Performance Metrics for DRL Agent vs APF} 
\begin{tabular}{|c|c|c|c|}
\hline
\textbf{Case (DRL)} & \textbf{Tracking ME (cm)} & \textbf{Tracking SD (cm)} & \textbf{TT (s)} \\
\hline
\textbf{1} & 4.80 & 0.2 & 5.00 \\
\hline
\textbf{2} & 4.30 & 0.3 & 4.50 \\
\hline
\textbf{3} & 5.50 & 0.3 & 4.20 \\
\hline
\textbf{4} & 5.50 & 0.3 & 6.40 \\
\hline
\textbf{5} & 5.60 & 0.2 & 5.20 \\
\hline
\end{tabular}
\begin{tabular}{|c|c|c|c|}
\hline
\textbf{Case (APF)} & \textbf{Tracking ME (cm)} & \textbf{Tracking SD (cm)} & \textbf{TT (s)} \\
\hline
\textbf{1} & NA & NA & NA \\
\hline
\textbf{2} & 15.6 & 1.0 & 14.7 \\
\hline
\textbf{3} & 14.2 & 0.5 & 14.8 \\
\hline
\textbf{4} & 13.4 & 1.1 & 15.5 \\
\hline
\textbf{5} & 16.0 & 1.2 & 14.0 \\
\hline
\end{tabular}
\label{mesdtt}
\vspace{-0.4cm}
\end{table}

Furthermore, Table \ref{mesdtt} indicates key performance parameters between the two approaches: mean error (ME), standard deviation (SD), and time taken (TT) to reach the target point for only successful flights. The DRL agent is able to perform high-precision gate and target tracking via velocity predictions, i.e., approximately 3.0 times better than the APF planner. Moreover, the time required to complete all the cases on average is 5.0 s, which is approximately 3.0 times faster than APF planning in dynamic environments. 

\begin{figure}[t]
\centering
\includegraphics[width=0.8\linewidth]{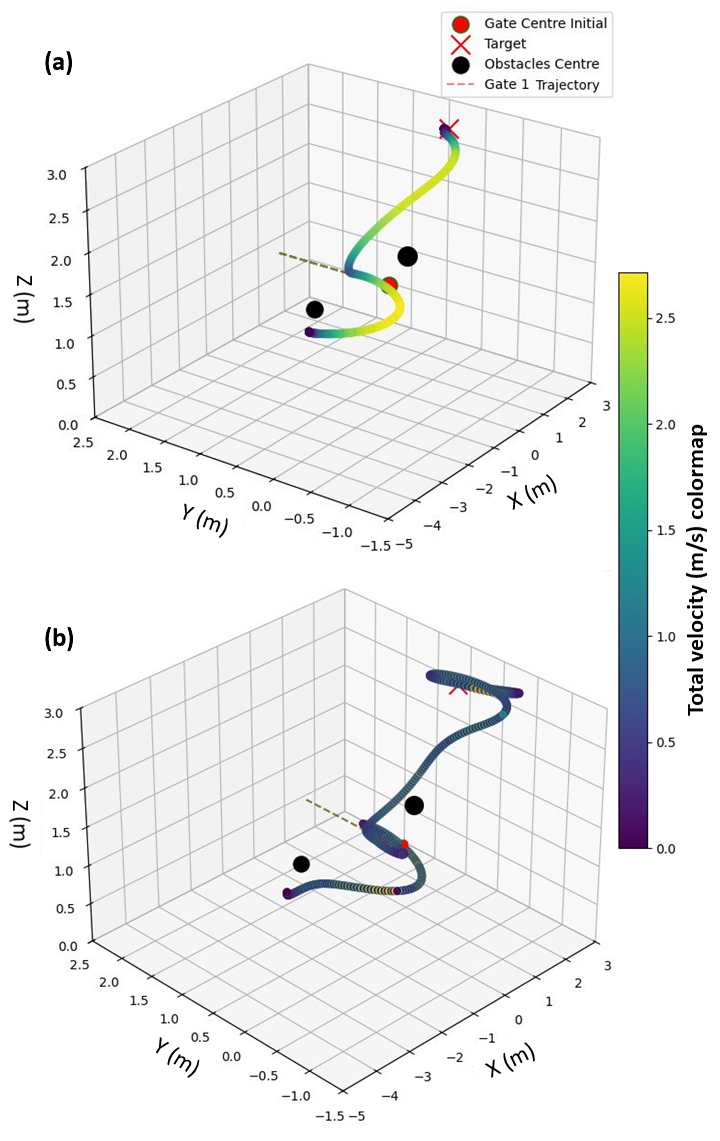} 
\caption{Trajectory of drones with velocity colormap of drone for Case 3 comparison where (a) shows trajectory using DRL agent and (b) shows trajectory via APF planning.}
\label{comparison}
\vspace{-0.3cm}
\end{figure}

From one of the visual representations of the cases, high precision with high-speed navigation is also evident from Fig. \ref{comparison} where it can be seen that the drone follows through the moving gate rather smoothly and intelligently, where its speed goes beyond 2.5 m/s. On the contrary, the APF planner struggles to adapt to the moving gate scenario and maintain its speed below 1.0 m/s in doing so.

\section{Experimental Evaluation}

\subsection{Experimental Setup}

\begin{figure}[t]
\centering
\includegraphics[width=0.8\linewidth]{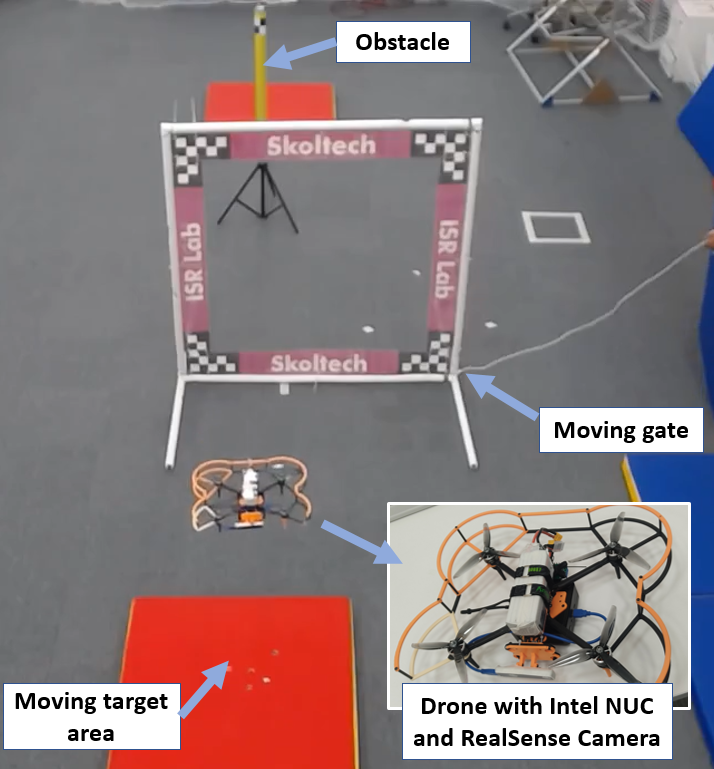} 
\caption{Experimental layout of AgilePilot: a drone equipped with RealSense camera performs dynamic navigation through the center of the gate while avoiding obstacles and reaching the target (red area).}
\label{exp}
\vspace{-0.3cm}
\end{figure}

Fig. \ref{exp} shows the general layout of our experimental setup. The drone, with a mass of 1.80 kg, is equipped with a high-performance Intel NUC onboard computer, alongside a RealSense T265 camera for real-time object detection. The onboard computer communicates with a SpeedyBee flight controller, which is responsible for sending and receiving commands through the ArduPilot firmware. For environmental layout, obstacles, gates, and landing pads are used where the objects are dynamically moved manually by a rope at speeds between 0.4 and 0.7 m/s to validate the transfer of simulation results to real-world conditions. 

The flight tests are conducted under both static and dynamic conditions. Three test cases were designed to assess the performance of the DRL agent, specifically focusing on its velocity predictions under varying conditions and its adaptability to the changing environment. The three cases are as follows:
\begin{itemize}
    \item \textbf{Case 1:} Slow left-moving gate with obstacle and target.
    \item \textbf{Case 2:} Fast right-moving gate with obstacle and target.
    \item \textbf{Case 3:} Static environment.
\end{itemize}

\subsection{Results} 

\subsubsection{Objects Position Estimation}

\begin{figure}[t]
\centering
\includegraphics[width=1.0\linewidth]{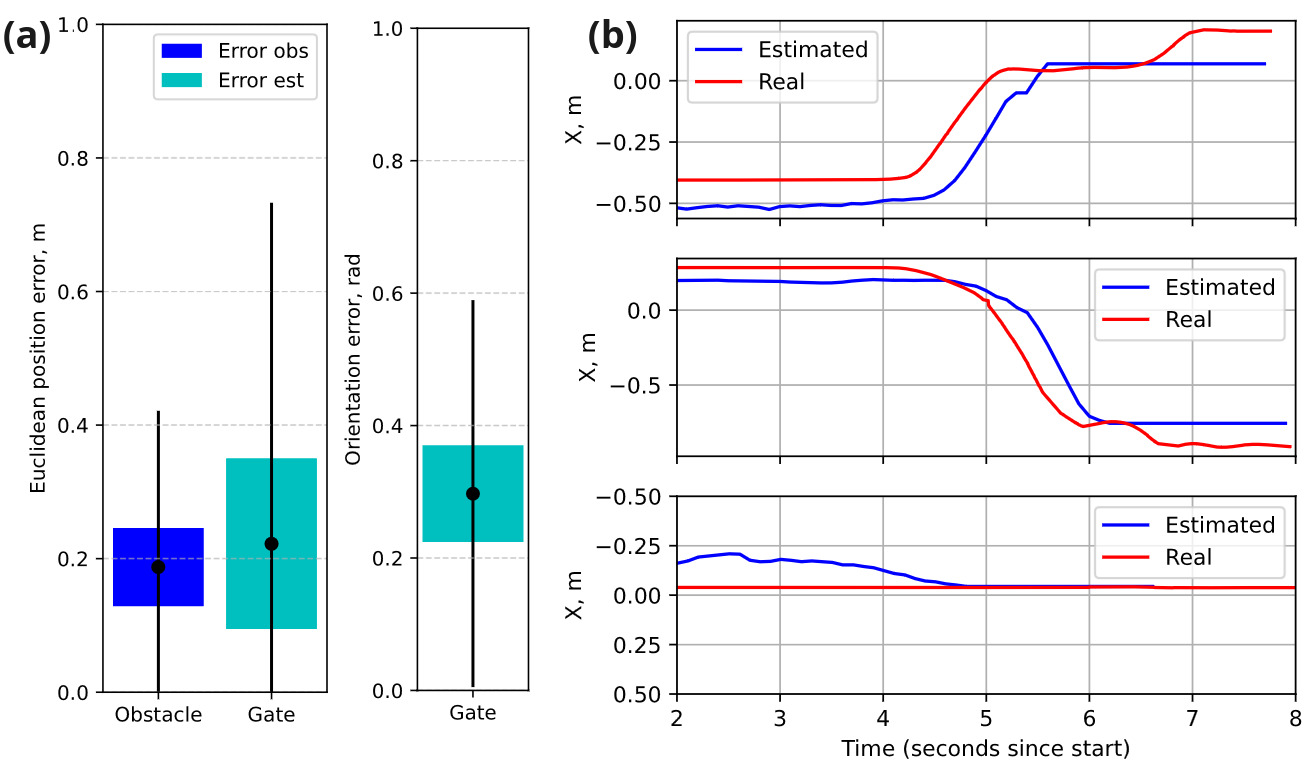} 
\caption{(a) Euclidean position and orientation errors for gate detection.
(b) Estimation of the X-coordinate position for moving gates for cases 1-3. The ground truth of the X coordinate (red line) and the predicted values (blue line).}
\label{errors}
\vspace{-0.5cm}
\end{figure}

The results of error measurements during flight experiments for the gate and obstacle pose estimation are shown in Fig. \ref{errors}. For position estimation, the mean position error is 0.19 m for obstacles and 0.22 m for gates. The standard deviation for position estimation is 0.06 m for obstacles and 0.13 m for gates. In terms of orientation, the mean orientation error for gates is 0.3 radians with a standard deviation of 0.07 radians, suggesting a reasonable accuracy in orientation estimation. The root mean square error (RMSE) for position is 0.076 m for obstacles and 0.37 m for gates. Additionally, a delay in the X-coordinate position change was observed (Fig. \ref{errors} b.), ranging up to 100 ms from the time when the real pose began to change to when the predicted X-coordinate pose started to respond.


\begin{figure*}[t]
\centering
\includegraphics[width=0.8\linewidth]{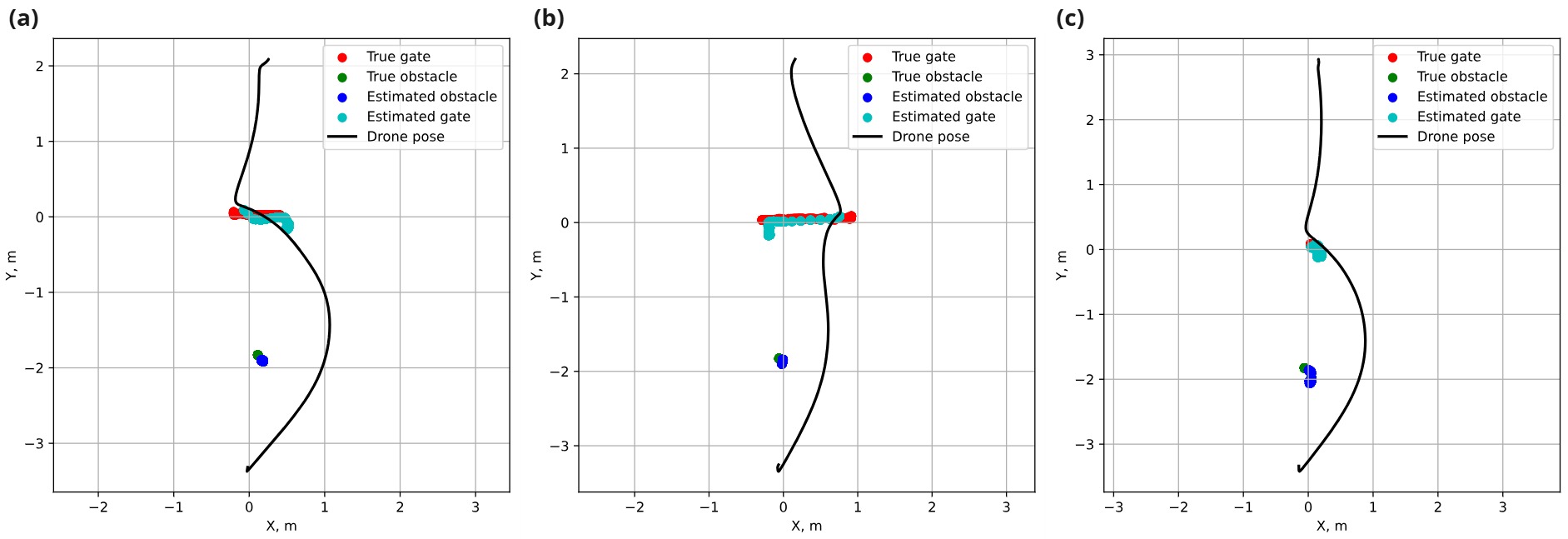} 
\caption{Top view of experiments: (a) trajectory of drone and detected poses of objects for Case 1, (b) detected poses of high-speed moving gate in Case 2 and (c) trajectory of drone and object detections for static environment in Case 3.}
\label{exp_pos}
\vspace{-0.3cm}
\end{figure*}

\begin{figure*}[t]
\centering
\includegraphics[width=0.8\linewidth]{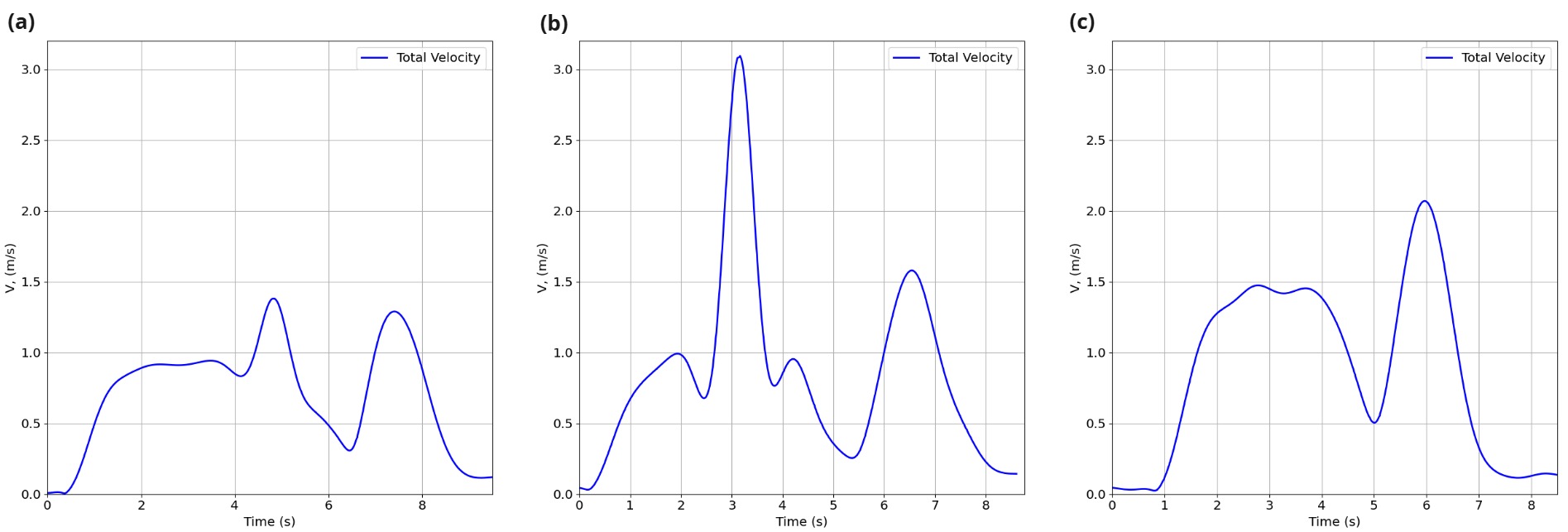} 
\caption{Total velocities: (a) Case 1, (b) Case 2, (c)  Case 3. The first peak in each case indicates drone movement when reaching the gate, and the second peak indicates drone movement towards the target point.}
\label{exp_vel}
\vspace{-0.3cm}
\end{figure*}

\subsubsection{Action predictions by DRL agent}

The position and velocity plots in Fig. \ref{exp_pos} and Fig. \ref{exp_vel} summarize that in Case 1 the drone was able to navigate without any abrupt changes through the gate being moved at a constant speed. The agent detects moving objects and gradually increases its speed to a maximum of 1.5 m/s while avoiding the obstacle smoothly. This smooth velocity profile suggests that the agent anticipated the gate's motion effectively.

In Case 2, the gate moved much faster as the drone approached it, presenting a highly dynamic scenario. To adjust, the drone increased its velocity up to 3.0 m/s. This quick adjustment also demonstrated the gate detection system's ability to accurately track the fast-moving gate. This behavior specifically underscores the robustness of the agent’s velocity management in dynamic conditions.

In Case 3, with the environment remaining static, the drone’s velocity predictions were smooth and consistent throughout the flight. There were no sudden changes in the environment, which allowed the drone to maintain a steady pace. After safely passing the gate and obstacle, the drone increased its speed, as there was no further risk of collision, leading to a more efficient and stable navigation path.

\section{Conclusion and Future work}

In conclusion, we presented AgilePilot, a novel DRL-based motion planner that enables intelligent navigation in dynamic environments by leveraging real-time object detection. The DRL framework was trained in dynamic simulation environments, facilitating a robust and adaptive learning process that bridges the simulation-to-reality gap effectively. The performance of our motion planner was compared with an APF-based motion planner in simulated dynamic environments. In summary, AgilePilot outperforms the classical approach by completing the task $3$ times faster and also $3$ times with more accuracy than the APF planner in terms of dynamic target tracking. Overall, our trained agent exhibited an average success rate of $90\%$ out of a total of 75 conducted experiments.

In the real world, three conditions were tested with varying environments and dynamic conditions by deploying the trained DRL agent onto hardware that utilized real-time object detection. The drone demonstrated safe navigation while avoiding obstacles and steadily passing through the dynamic gate. However, when the environment was subject to high-speed changes, such as rapidly moving gates, the drone displayed remarkable agility and adaptability, adjusting its speed up to 3 m/s to meet the dynamic demands of the environment. The real-time detection of gates and obstacles was accurate, encompassing an RMSE of 0.076 m for obstacles and 0.37 m for the moving gates. The observed delay results from the time taken for image acquisition, neural network processing, and the transformation of position and pose filtering. However, this delay does not significantly impact performance, and the drone performs well in experiments. These results validate the effectiveness of AgilePilot in real-world conditions, demonstrating its capability to navigate dynamic and unpredictable environments using real-time velocity predictions from trained agents.

In the future, more complex scenarios could be explored, including the introduction of unpredictable obstacles and environments with multiple agents. Additionally, further research could focus on fully autonomous systems that utilize CV not only for object detection but also for precise drone localization, eliminating the need for external position systems. This would improve situational awareness and expand the range of deployable scenarios for autonomous drone systems.

\section*{Acknowledgements} 
Research reported in this publication was financially supported by the RSF grant No. 24-41-02039.

\bibliographystyle{IEEEtran}
\bibliography{bibliography}
\balance
\addtolength{\textheight}{-12cm}
\end{document}